\documentclass[10pt,twocolumn,letterpaper]{article}

\usepackage{wacv}                       

\usepackage{times}
\usepackage[T1]{fontenc}
\usepackage{epsfig}
\usepackage{graphicx}
\usepackage{amsmath}
\usepackage{amssymb}
\usepackage[accsupp]{axessibility}  
\usepackage{booktabs}
\usepackage{multirow}
\usepackage{array}
\usepackage{xcolor}
\usepackage{enumitem}
\usepackage{algorithm}
\usepackage{algorithmic}
\usepackage{dblfloatfix}
\definecolor{wacvblue}{rgb}{0.21,0.49,0.74}
\usepackage[pagebackref,breaklinks,colorlinks,allcolors=wacvblue]{hyperref}

\def\wacvPaperID{190}
\def\confName{WACV}
\def\confYear{2027}

\title{Beyond Region Memory: Local Point Tracking for Vocal Cord Localization in Video Laryngoscopy}

\author{
Huayu Wang$^{1}$ \quad Gyungmin Ko$^{1}$ \quad Jung Heon Kim$^{1,2}$ \quad Jia-Xian Jian$^{3}$\\
Dae Kon Kim$^{4}$ \quad Dong Keon Lee$^{5}$ \quad Hsiang-Wei Huang$^{1}$ \quad Pau-Choo Chung$^{3}$\\
Linda Shapiro$^{1}$ \quad Richard B. Utarnachitt$^{1,6}$ \quad Jenq-Neng Hwang$^{1}$ \quad Nathan J. White$^{1,6}$\\[4pt]
{\small $^{1}$University of Washington \quad $^{2}$Ajou University \quad $^{3}$National Cheng Kung University}\\
{\small $^{4}$Gangneung Asan Hospital \quad $^{5}$Seoul National University Bundang Hospital \quad $^{6}$Airlift Northwest}
}

\begin{document}
\maketitle

\begin{abstract}
Reliable vocal-cord localization in emergency video laryngoscopy remains visually challenging due to rapid tissue deformation, specular highlights, and instrument occlusions. We propose \emph{Local Point Tracking} (LPT), a robust framework that combines a single-frame detector, local point propagation, and stateless mask refinement. LPT follows a \emph{localize first, snap second} pipeline. First, a single-frame semantic detector generates sparse anchors, within which LPT initializes local query points (SAM-supported by default, with a regular grid as fallback). A point-propagation module tracks these points through detector-missed frames and recovers temporal bounding boxes from their geometry. Second, because point-derived boxes can be loose around the visible boundary, a box-snap step refines the selected box to the target boundary. This decomposition keeps temporal state in local point coordinates rather than a stateful mask or whole-region representation, while the spatial refinement module handles single-frame boundary precision without feeding mask errors back into temporal memory. Extensive evaluations on two distinct datasets demonstrate that our proposed LPT method achieves strong overall performance against existing baselines, with the best mAP$_{50}$, success AUC, and missing rate on both datasets. Furthermore, ablation studies confirm that point tracking primarily enhances temporal tracking robustness, while box-snap refinement effectively improves boundary tightness.
\end{abstract}

\section{Introduction}
\begin{figure*}[!t]
    \centering
    \includegraphics[width=\textwidth]{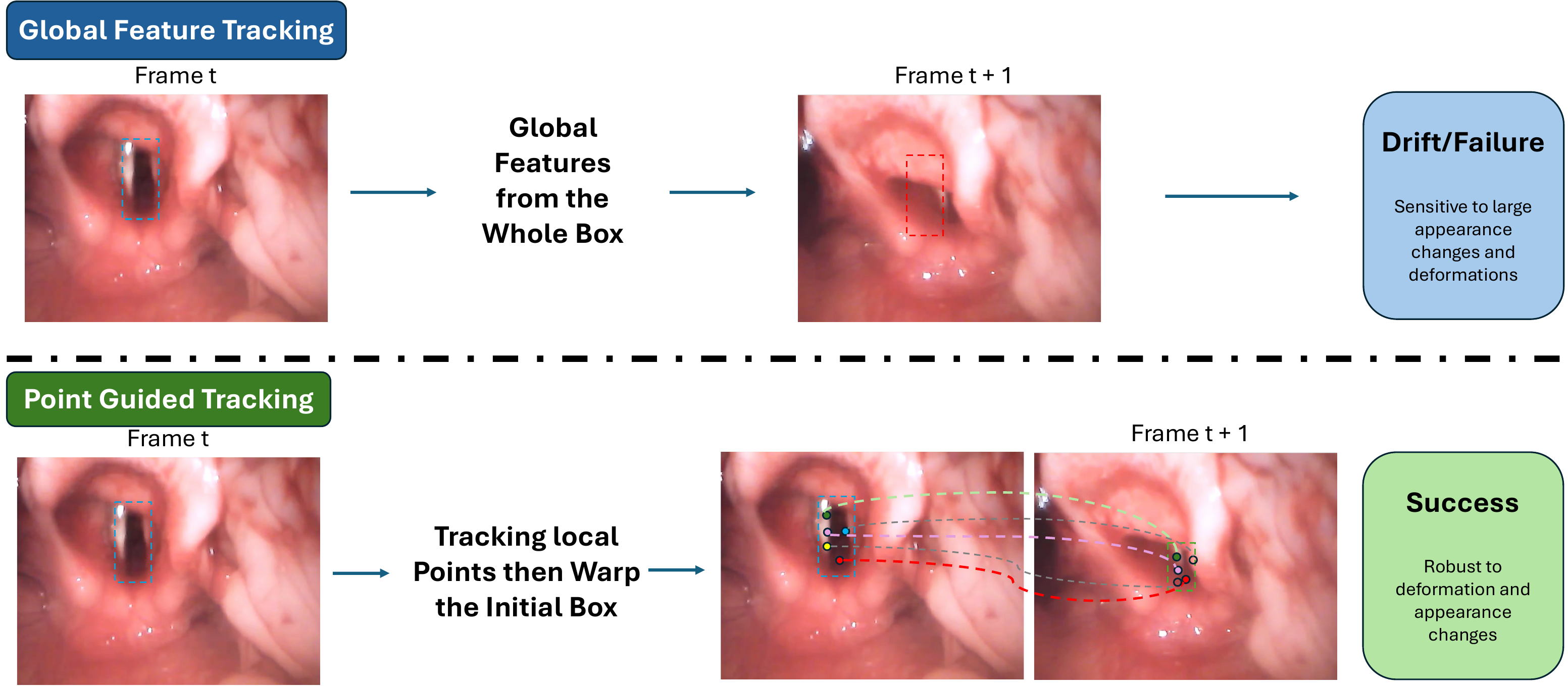}
    \caption{
    \textbf{Motivation of local point tracking.}
    Region-level trackers propagate whole-box or mask features and may drift under deformation, occlusion, specular glare, or similar surrounding tissue.
    LPT instead tracks target-supported local points, recovers a box from their motion, and applies mask-snap refinement for tighter boundaries.
    }
    \label{fig:global_vs_point_tracking}
\end{figure*}

Video laryngoscopy is widely used during emergency endotracheal intubation to visualize the airway and guide tube placement. Automated assistance during this procedure depends strongly on accurate localization of the vocal cords or glottic opening\cite{choi2023mask}. This localization is an important prerequisite for downstream tasks such as glottal-area analysis, view-quality feedback, and procedure-state recognition\cite{andrade2025girafe}, and ultimately for real-time navigation assistance and semi-autonomous intubation tools. Yet, this setting is visually difficult: the camera is handheld, the anatomy deforms rapidly, the target may be partially occluded by instruments or secretions, and specular highlights often dominate the image\cite{lossius2012patient}. Reliable localization therefore requires both semantic grounding and temporal recovery\cite{kim2023development}.

Single-frame detectors provide strong semantic grounding~\cite{carion2020end}. A fine-tuned single-frame detection model can identify the target in clear frames, but frame-wise detection has no temporal memory and can miss blurred or occluded frames. Video trackers and video segmentation foundation models address the complementary problem by propagating target state over time. Detection-integrated trackers such as ByteTrack~\cite{zhang2022bytetrack}, BoT-SORT~\cite{aharon2022bot}, and CL-MC~\cite{wang2026detector}, which uses detections to correct the memory of Segment Anything Model 2 (SAM2)~\cite{ravi2025sam}, further improve robustness, but they often maintain a mask-level or region-level state through the sequence. In emergency laryngoscopy, the target boundary is non-rigid, the surrounding tissue has similar color and texture, and distractors such as glare, tools, and occlusions can occupy a substantial portion of the tracked region. A broad region state can therefore absorb distracting context, especially during visually ambiguous frames.

Our model starts by making the temporal representation more local. Instead of tracking the whole region, it tracks target-supported points initialized from detector anchors. At confident anchors provided by a single-frame detector (e.g., YOLO\cite{tian2026yolov12}), SAM2 is used as a single-frame image module to sample query points inside the detected target. A pre-trained point-tracking algorithm (CoTracker\cite{karaev2025cotracker3}) then propagates these points through detector-missed frames. Because points are local correspondences, they can follow deformable motion without requiring a single global region feature to remain clean. LPT recovers a box from the propagated points with a simple robust point-to-box module that combines a visibility-weighted affine warp and a statistical quantile envelope, and ranks detector and point-tracking candidates with a deterministic candidate-scoring rule.

Robust temporal discovery is not the same as strict spatial tightness. The point-derived box tells us where the vocal cords are, but it may be loose around the visible object extent. We therefore add a stateless refinement stage: the selected box is passed back to SAM2 as a single-frame box prompt, and the resulting mask is converted into a tight bounding box. This yields a \emph{localize first, snap second} design. By orchestrating these pre-trained models, the detector and point tracker provide temporal localization with reduced drift, while SAM2 is used only for single-frame point initialization and final boundary snapping, not for long-term video memory. This gives a practical alternative to carrying a full region or mask state across difficult frames. With LPT, we study whether a SAM-supported local point state, followed by stateless single-frame refinement, provides a more stable operating point for this clinical localization task.

In summary, rather than introducing a monolithic tracker, LPT provides a modular framework for vocal-cord localization. It uses target-supported point trajectories to reduce exposure to whole-region memory errors in highly ambiguous surgical scenes. The point-to-box recovery and selector modules are deliberately simple; ablations show that LPT is insensitive to the specific point-sampling and box-recovery geometry, suggesting that the main gains come from point-based temporal gap recovery and the final box snap rather than a finely tuned geometric estimator. We further incorporate a stateless box-snap refinement stage that improves boundary tightness without feeding mask errors back into a temporal memory loop. Comprehensive evaluations on emergency laryngoscopy videos, including two datasets from different hospitals, empirically validate this design: the point-tracking mechanism effectively drives temporal recovery, while the stateless refinement improves spatial precision.

\section{Related Work}

\paragraph{Tracking-by-Detection Paradigms.}
Visual tracking has long been dominated by the tracking-by-detection paradigm\cite{bewley2016simple,wojke2017simple,cao2023observation}, which connects frame-wise bounding boxes into continuous trajectories. Modern trackers such as ByteTrack~\cite{zhang2022bytetrack} and BoT-SORT~\cite{aharon2022bot} have achieved strong robustness in this domain. However, in surgical environments like emergency laryngoscopy, spatial tightness and temporal continuity can be difficult to maintain simultaneously. A fine-tuned single-frame detector provides highly accurate and tight bounding boxes on clear frames, but it frequently fails under motion blur, specular glare, or instrument occlusion. While bounding-box trackers bridge these temporal gaps by associating detections across time, they propagate a holistic region representation that struggles with non-rigid target deformation~\cite{sun2026understanding}. Consequently, their tracked boxes often drift or become excessively loose, absorbing surrounding tissue context and failing to maintain the strict spatial precision required for clinical downstream tasks.

\paragraph{Point Tracking and Propagation.}
To handle non-rigid deformation and fine-grained correspondence, point-based tracking offers a flexible representation\cite{doersch2023tapir, doersch2024bootstap, harley2022particle, karaev2024cotracker}. Algorithms like CoTracker~\cite{karaev2025cotracker3} jointly track a grid of points across videos and can model local physical motion without committing to a single global box or mask state. Building on this, methods like SAM-PT~\cite{rajivc2025segment} combine sparse point trajectories with the Segment Anything Model (SAM)\cite{kirillov2023segment}. However, continuously decoding a mask from propagated points can still be sensitive to point drift. LPT instead derives bounding boxes directly from point geometry through visibility-weighted affine and quantile estimation, and then uses SAM2 as a stateless, single-frame boundary snap to recover the tightness lost during temporal propagation.

\paragraph{Foundation Models and Detector-in-the-Loop Tracking.}
The advent of vision foundation models has reshaped video segmentation and tracking. Recent zero-shot trackers like SAMURAI~\cite{yang2026samurai} adapt SAM2 for motion-aware tracking, while clinical detector-in-the-loop systems like CL-MC~\cite{wang2026detector} use a fine-tuned detector to intermittently reset SAM2's stateful video memory\cite{cheng2022xmem, cheng2024putting}. These methods are strong baselines, but in visually hostile surgical scenes a propagated mask or region feature can absorb distracting background and drift. LPT uses a different decomposition: local point coordinates carry the temporal state, while the segmentation foundation model is invoked only for single-frame point initialization and stateless spatial refinement.

\section{Method}

LPT localizes the vocal-cord/glottic region by separating temporal localization from spatial boundary estimation. The pipeline follows a \emph{localize first, snap second} cascade, integrating a single-frame semantic detector, a point-tracking algorithm, and a foundation segmentation model. 

The process begins with the single-frame detector, which supplies high-confidence semantic anchors across the video. At any given anchor frame, SAM2 is used as a single-frame image module to sample target-supported query points within the detected bounding box. Next, CoTracker propagates these local points through subsequent frames. Finally, SAM2 is invoked again as a stateless refiner, snapping the point-derived temporal box to the visible anatomical boundary. By composing these models without introducing a learned gating network, the framework reduces reliance on long-term global region features in complex surgical environments.

\begin{figure*}[!tb]
    \centering
    \includegraphics[width=0.88\textwidth]{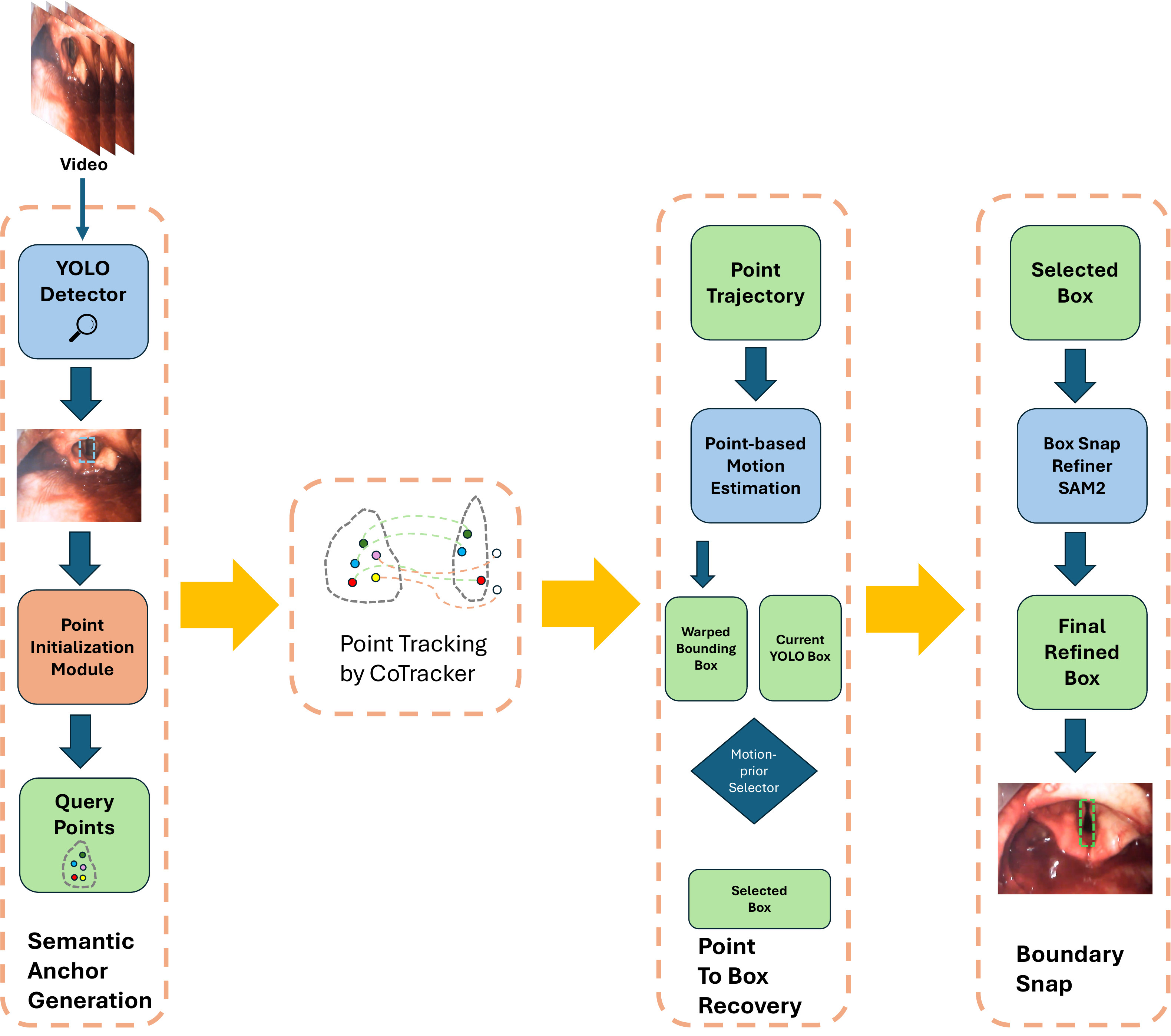}
    \caption{
    \textbf{Overview of LPT.}
    The pipeline follows a \emph{localize first, snap second} cascade.
    A YOLO anchor initializes target-supported points, CoTracker propagates them, point-to-box recovery and motion-guided scoring select a candidate box, and a stateless SAM2 box-snap refiner produces the final tight box.
    Point initialization, propagation, and box recovery are described in Sec.~\ref{subsec:local_point_tracking}; candidate scoring and box-snap refinement are described in Sec.~\ref{subsec:selection_refine}.
    Blue, orange, and green blocks denote learned models, deterministic modules, and intermediate or output boxes.
    }
    \label{fig:architecture}
\end{figure*}

\subsection{Local Point Tracking}
\label{subsec:local_point_tracking}

Let $I_t$ denote the input frame and let $\mathcal{D}(I_t)=\{d_{t,k}\}$ be the detector outputs. A new tracking segment starts from a confident detector box that is sufficiently different from the previous anchor. Denote the anchor box by $b_0$ and the sampled query points by $q_i\in b_0$, $i=1,\ldots,N$. In the default implementation, SAM2 provides the sampling support through a single-frame box prompt; if the prompt mask is empty, we sample a regular grid inside $b_0$. CoTracker then propagates the query set and returns point positions $p_{t,i}$ with a binary visibility mask $v_{t,i}\in\{0,1\}$.

To convert local correspondences back to a robust bounding box, we estimate a visibility-weighted affine transform from the anchor points to the current points:
\begin{equation}
\begin{aligned}
A_t^\star &= \arg\min_A \sum_i w_{t,i}\|A\bar q_i - p_{t,i}\|_2^2,\\
w_{t,i} &= v_{t,i} \exp\!\left(-\|q_i-c_0\|_2^2 / 2\sigma_0^2\right).
\end{aligned}
\end{equation}
Here $\bar q_i$ is the homogeneous point coordinate, $c_0$ is the anchor-box center, and $\sigma_0$ is set from the mean distance between the initial query points and the anchor-box center. Warping $b_0$ with $A_t^\star$ gives an affine box $b_t^{\mathrm{aff}}$. If the weighted affine fit is ill-conditioned, we fall back to a RANSAC similarity fit\cite{fischler1981random}. In parallel, we compute a robust quantile box $b_t^{\mathrm{q}}$ from the visible points, using the RANSAC inlier subset when the fallback fit is used. The overall track reliability is the visible-point fraction:
\begin{equation}
r_t=\frac{1}{N}\sum_i v_{t,i}.
\end{equation}
The track-derived candidate box is then dynamically blended based on this reliability:
\begin{equation}
b_t^{\mathrm{trk}} = r_t b_t^{\mathrm{aff}} + (1-r_t)b_t^{\mathrm{q}} .
\end{equation}
This simple fusion provides a stable default that balances smooth tissue motion captured by the affine warp with the statistical robustness of a point envelope. As shown in the ablation, LPT is not highly sensitive to the exact recovery estimator.

\subsection{Selection and Box Snap}
\label{subsec:selection_refine}

At each frame, we may have detector boxes, the track-derived box, or both. To choose an output without training a learned gate, LPT uses a deterministic, motion-guided selector. The track-derived box defines a motion-prior center $\mu_t$, and the visible tracked points define a covariance $\Sigma_t$ with a box-size-dependent diagonal regularizer. For a detector box $d_{t,k}$ with confidence $c_{t,k}$ and center $z_{t,k}$, we first compute its agreement with the point-track prior:
\begin{equation}
g_{t,k} =
\exp\!\left[
-\frac{1}{2}(z_{t,k}-\mu_t)^\top
\Sigma_t^{-1}(z_{t,k}-\mu_t)
\right].
\end{equation}
The default implementation then uses a deterministic soft-scored argmax:
\begin{equation}
\begin{aligned}
S_{\mathrm{det}}(d_{t,k}) &= c_{t,k}\left(1+\lambda r_t g_{t,k}\right),\\
S_{\mathrm{trk}} &= \beta r_t^2 ,
\end{aligned}
\end{equation}
where $\lambda$ and $\beta$ are fixed constants (all hyper-parameter values, including $N$, are listed in the supplementary material). Detector confidence and point visibility are not assumed to be calibrated probabilities: the detector score stays anchored to YOLO confidence with a bonus for agreeing with the tracked-point prior, while the track score is discounted by its visible-point fraction $r_t$. The pre-refinement output $\tilde b_t$ is the highest-scoring candidate. This selector is therefore best viewed as a simple confidence-ranking heuristic that combines detector semantics with point-track consistency, rather than as a learned probabilistic gate.
The selected box $\tilde b_t$ robustly localizes the target, but point-derived geometry can be loose around the visible boundary. Therefore, the pipeline executes a final stateless box snap: SAM2 is prompted with $\tilde b_t$ on the current frame, and the resulting segmentation mask is converted into its tight enclosing bounding box, which is emitted as the final prediction. If the generated mask is empty or the enclosing box is invalid after image-boundary clipping, the system defaults to $\tilde b_t$. This final step delegates strict spatial precision to the foundation model without allowing potential mask errors to corrupt any underlying temporal memory.

\section{Experiments}

\begin{table*}[!tb]
\centering
\caption{Main detection results. Harborview follows the CL-MC benchmark protocol; SNUBH reports all 23 sequences. CL-MC is tested with the same YOLO weights used by detector-dependent methods. LPT (online) is the streaming variant described in Sec.~\ref{sec:setup}. \textbf{Bold} = best, \underline{underline} = second.}
\label{tab:main}
\scriptsize
\setlength{\tabcolsep}{4pt}
\renewcommand{\arraystretch}{0.9}
\resizebox{\linewidth}{!}{%
\begin{tabular}{l cccc cccc}
\toprule
 & \multicolumn{4}{c}{\textsc{Harborview}} & \multicolumn{4}{c}{\textsc{SNUBH} (23 seq.)} \\
\cmidrule(lr){2-5}\cmidrule(lr){6-9}
Method & mAP$_{50}\uparrow$ & mAP$_{50:95}\uparrow$ & AUC$\uparrow$ & MR$\downarrow$
       & mAP$_{50}\uparrow$ & mAP$_{50:95}\uparrow$ & AUC$\uparrow$ & MR$\downarrow$ \\
\midrule
YOLOv12                   & 81.65 & 54.68 & 74.84 & 10.58 & 54.20 & 29.27 & 47.27 & 41.88 \\
ByteTrack          & 81.16 & 50.52 & 73.10 & 10.92 & 53.28 & 26.61 & 46.06 & 42.59 \\
BoT-SORT            & 81.84 & 54.08 & 74.57 & 10.39 & 54.24 & 28.65 & 46.91 & 41.85 \\
CL-MC          & 84.32 & 50.95 & 76.52 & 6.85 & 56.50 & 32.67 & 55.53 & 35.25 \\
SAM-PT                 & 68.82 & 48.80 & 67.27 & 23.35 & 40.07 & 29.43 & 41.96 & 54.62 \\
SAMURAI               & 75.37 & 52.73 & 70.89 & 20.25 & 58.21 & \textbf{44.37} & 59.07 & 37.20 \\
\midrule
\textbf{LPT} (w/o box snap)       & \underline{88.32} & 57.04 & 78.48 & \underline{6.36} & 59.96 & 30.31 & 54.91 & \underline{34.13} \\
\textbf{LPT} (default)        & \textbf{89.06} & \textbf{67.87} & \textbf{84.06} & \textbf{5.57} & \textbf{61.02} & 42.52 & \textbf{61.61} & \textbf{33.12} \\
\textbf{LPT} (online)     & 87.65 & \underline{67.74} & \underline{83.41} & 6.45 & \underline{60.02} & \underline{42.66} & \underline{59.88} & 34.66 \\
\bottomrule
\end{tabular}%
}
\end{table*}

\subsection{Experimental Setup}
\label{sec:setup}

\paragraph{Detector and model components.}
To isolate the effect of temporal localization, all detector-dependent methods use the same single-frame detector. Following CL-MC, we use a YOLOv12-m detector trained on public non-emergency laryngoscopy data, including Laryngoscope8\cite{yin2021laryngoscope8} and the additional public images used by the prior detector\cite{tian2026yolov12}. We test CL-MC with the same YOLO weights used by YOLO-only, tracking-by-detection baselines, and LPT. LPT then composes this detector with SAM2 for stateless point initialization, CoTracker for local point propagation, and SAM2 again for box-snap refinement. Unless otherwise stated, ``LPT'' denotes the default offline local-tracking method with SAM-mask-supported point initialization and the final box snap; ``LPT (w/o box snap)'' removes only the last refinement step. ``LPT (online)'' advances CoTracker in 8-frame steps and emits buffered frames incrementally, without revising earlier outputs. The offline default instead uses the final trajectories of each complete tracking segment. Both use the same re-anchoring rule; the online runner flushes pending frames before re-anchoring and at video end, and applies box snap synchronously to each emitted frame. The 8-frame step specifies the output schedule, not the processing throughput.

\paragraph{Evaluation datasets.}
We evaluate on two emergency video-laryngoscopy datasets from different clinical centers. \textbf{Harborview} is the benchmark used by CL-MC, containing 24 emergency intubation videos collected during prehospital air medical transport. There are 8,931 frames (297 seconds) with frame-level annotations. The videos contain rapid camera motion, secretions, specular highlights, and instrument occlusion, and are disjoint from the detector training data. We use the same test split and annotation protocol as CL-MC for direct comparison. \textbf{SNUBH} is an independent 23-video emergency-laryngoscopy set annotated at frame level. There are 13,230 frames, 441 seconds in total. Following the annotation style of prior vocal-cord detection work, the visible vocal-cord/glottic region is marked with polygon annotations and converted to axis-aligned bounding boxes for detection evaluation. 

\paragraph{Metrics and baselines.}
We report mAP$_{50}$, the average precision of confidence-ranked predicted boxes at an IoU threshold of 0.5, and mAP$_{50:95}$, the average precision averaged over IoU thresholds from 0.50 to 0.95 in steps of 0.05~\cite{lin2014microsoft}. Success AUC~\cite{wu2013online} averages the success rate over 101 IoU thresholds, $0,0.01,\ldots,1$. At each threshold, success is the fraction of ground-truth frames whose IoU exceeds it; missing predictions have IoU zero. Missing Rate (MR) is the fraction of ground-truth frames with no prediction or with IoU below 0.5. Each metric is computed per video and then averaged with equal video weights; longer videos do not receive greater weight in the dataset-level scores. mAP$_{50}$ and MR measure target discovery and temporal recovery, while mAP$_{50:95}$ stresses boundary tightness. We compare against the single-frame YOLO detector, tracking-by-detection baselines ByteTrack and BoT-SORT, CL-MC, and foundation-model trackers SAM-PT and SAMURAI. For SAM-PT and SAMURAI, the tracker is initialized from the first high-confidence detector box and then propagated forward. Unlike the detector-in-the-loop methods (CL-MC and LPT), these two trackers are not re-anchored to the detector after initialization; we isolate the effect of detector re-anchoring in the ablation of Table~\ref{tab:abl_modules}. The overall comparison is summarized in Table~\ref{tab:main}, with detector-gap recovery, external deformable-object generalization, and efficiency analyzed separately in Tables~\ref{tab:gap_recovery}, \ref{tab:external_cholecseg8k}, and~\ref{tab:runtime}.

\paragraph{External deformable-object benchmark.}
To evaluate whether the proposed localization-and-refinement design generalizes to other deformable surgical targets, we additionally test on CholecSeg8k\cite{hong2020cholecseg8k}, a public laparoscopic cholecystectomy semantic-segmentation dataset derived from Cholec80\cite{twinanda2016endonet}. CholecSeg8k contains soft-tissue and instrument classes with substantial shape variation, including liver, abdominal wall, gallbladder, fat, grasper, and l-hook. Since the dataset provides semantic masks rather than box annotations, we convert each evaluated class mask to its enclosing box and report class-wise AP$_{50}$ for six surgical-scene classes. This box-level protocol evaluates localization only and discards the boundary detail available in the masks.

\subsection{Main Results}

The main comparison in Table~\ref{tab:main} evaluates detector-only, tracking-by-detection, and foundation-model tracking baselines under the same metric set. On the Harborview benchmark, LPT achieves the best performance across all reported metrics in this table, improving over the CL-MC baseline by 4.74 mAP$_{50}$, 16.92 mAP$_{50:95}$, and 7.54 AUC points, while reducing MR from 6.85 to 5.57. The largest gain appears on the strict mAP$_{50:95}$ metric, supporting the role of the stateless box-snap module after temporal localization.

On the independent SNUBH dataset, LPT similarly achieves the best mAP$_{50}$, AUC, and MR. Among the non-proposed baselines, SAMURAI obtains the highest strict mAP$_{50:95}$, but its MR is higher (37.20 vs. LPT's 33.12). This indicates that stateful mask propagation can produce tight boxes when it remains locked on the target, yet it is less reliable across detector gaps and ambiguous frames. In contrast, the local point state reduces missing frames, and the subsequent box snap recovers much of the boundary tightness. The online variant remains close to the offline default, suggesting that rolling-window inference preserves the same accuracy trend while enabling streaming deployment.

\subsection{External Deformable-Object Generalization}
\label{sec:exp_external}

\begin{table}[!htb]
\centering
\caption{External public-dataset stress test on CholecSeg8k. We report class-wise AP$_{50}$ in percent for six evaluated classes after converting semantic masks to box targets under a fixed protocol. LPT denotes the default method with SAM2 box-snap refinement. This experiment is not used for tuning. \textbf{Bold} = best.}
\label{tab:external_cholecseg8k}
\scriptsize
\setlength{\tabcolsep}{3.5pt}
\renewcommand{\arraystretch}{0.95}
\resizebox{\linewidth}{!}{%
\begin{tabular}{l|cccccc}
\toprule
Class & \textbf{LPT} & YOLO-only & ByteTrack & CL-MC(base) & SAM-PT & SAMURAI \\
\midrule
liver          & \textbf{81.58} & 70.10 & 34.79 & 75.45 & 66.88 & 77.22 \\
abdominal wall & 74.71 & 68.03 & 53.96 & \textbf{75.93} & 48.39 & 69.34 \\
grasper        & 72.64 & 52.74 & 37.16 & \textbf{75.23} & 46.53 & 43.33 \\
gallbladder    & \textbf{54.97} & 24.48 & 14.49 & 48.59 & 46.43 & 21.87 \\
fat            & 48.77 & 43.15 & 21.18 & \textbf{49.43} & 31.81 & 23.47 \\
l-hook         & \textbf{29.20} & 27.82 & 16.29 & 28.54 & 12.68 & 26.10 \\
\midrule
Mean (Rigid)   & 50.92 & 40.28 & 26.73 & \textbf{51.89} & 29.61 & 34.72 \\
Mean (Deform)  & \textbf{65.01} & 51.44 & 31.11 & 62.35 & 48.38 & 47.98 \\
\textbf{Mean (All)} & \textbf{60.31} & 47.72 & 29.65 & 58.86 & 42.12 & 43.56 \\
\bottomrule
\end{tabular}%
}
\end{table}

Table~\ref{tab:external_cholecseg8k} evaluates whether the same localization-and-refinement design transfers outside the vocal-cord domain to surgical-scene objects with different motion and shape properties. The results show an advantage on deformable soft-tissue structures: LPT obtains the best Mean (Deform) AP$_{50}$, improving over YOLO-only by 13.57 points and over CL-MC by 2.66 points. This pattern is especially visible on liver and gallbladder, where local point trajectories can follow non-rigid shape changes before the stateless box snap tightens the final localization. On rigid instrument classes, CL-MC is slightly stronger on average, but LPT achieves the best all-class mean. Here we group the four soft-tissue classes (liver, abdominal wall, gallbladder, fat) as \emph{deformable} and the two instrument classes (grasper, l-hook) as \emph{rigid}.

\begin{figure*}[!tb]
    \centering
    \includegraphics[width=\textwidth]{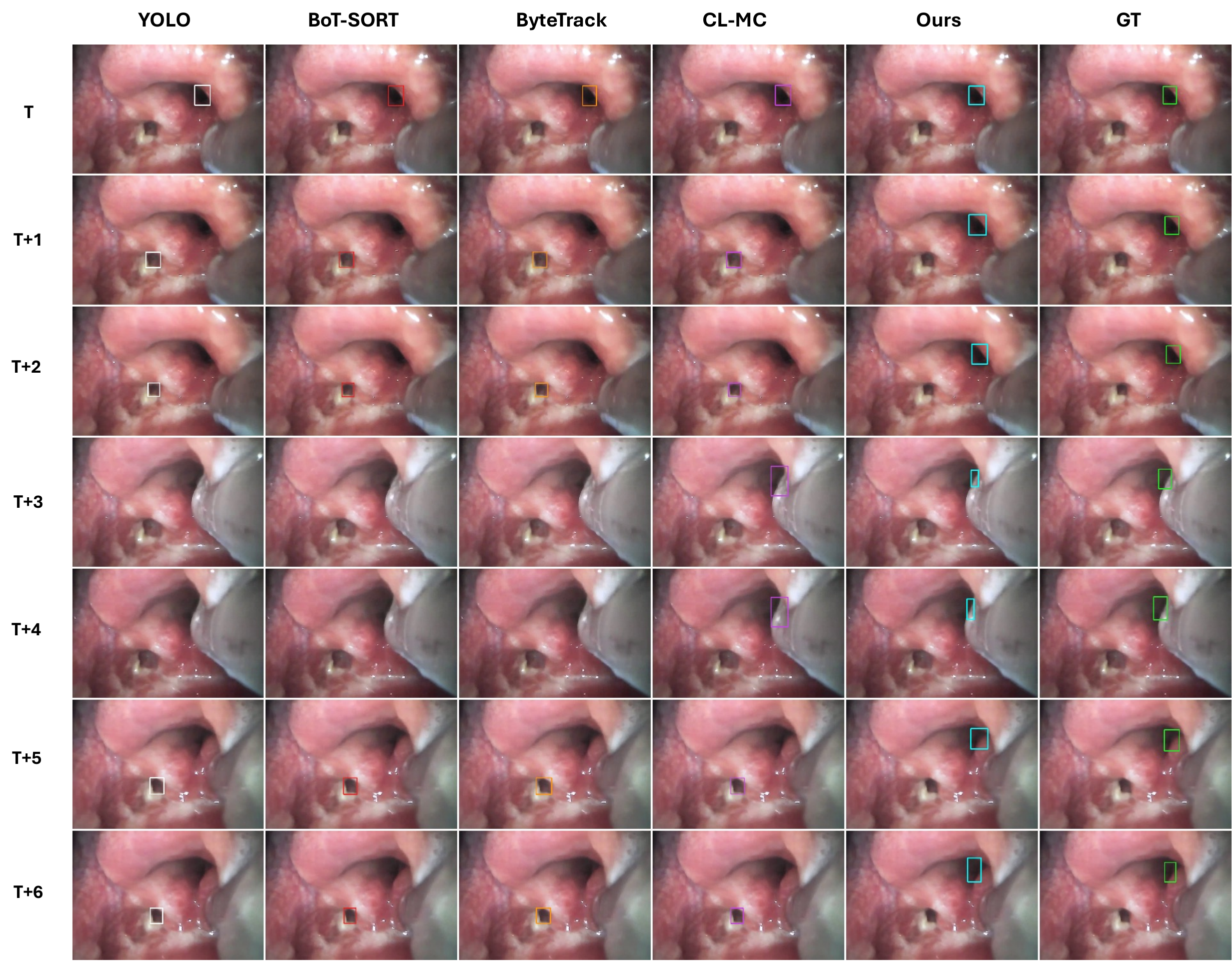}
    \caption{
    \textbf{Qualitative comparison under detector failure.}
    Rows show sampled frames from one sequence (time increases downward; gaps are non-uniform), and columns show each method's predicted box:
    \textcolor{gray}{YOLO} (white),
    \textcolor{red}{BoT-SORT} (red),
    \textcolor{orange}{ByteTrack} (orange),
    \textcolor{magenta}{CL-MC} (purple),
    \textbf{LPT} (cyan), and the ground truth (green).
    After an initially correct detection, YOLO becomes unreliable under glare, deformation, and rapid motion, causing detection-dependent trackers to drift or drop the target.
    LPT maintains localization with local point trajectories and then snaps the final box to the visible boundary.
    }
    \label{fig:drift_vs_continuous_tracking}
\end{figure*}

The qualitative example in Figure~\ref{fig:drift_vs_continuous_tracking} illustrates the same failure mode targeted by LPT. When the single-frame detector becomes unreliable under specular glare, rapid tissue deformation, or fast camera motion, tracking-by-detection baselines and detector-in-the-loop trackers can propagate the misleading region into their state, leading to drift or dropped predictions. In contrast, because LPT delegates the temporal state to local point correspondences rather than a global mask, it remains centered on the vocal cords through this difficult window and can recover a tight box once the stateless refinement is applied.

\subsection{Recovery in Detector-Missed Intervals}
\label{sec:exp_gap_recovery}

\begin{table}[!htb]
\centering
\caption{\textbf{Recovery under detector-missed intervals.} Gaps are maximal labelled-frame runs where YOLO has IoU$<$0.5. Recovery is the percentage of gap frames re-localized at IoU$\ge$0.5; Mean IoU is the IoU of LPT (default) averaged over all gap frames in the bucket. \textbf{Bold} = best recovery.}
\label{tab:gap_recovery}
\scriptsize
\setlength{\tabcolsep}{4pt}
\renewcommand{\arraystretch}{0.9}
\resizebox{\linewidth}{!}{%
\begin{tabular}{l l r ccc c}
\toprule
Dataset & Gap len. & \#gaps & CL-MC & LPT\,(w/o box snap) & LPT\,(default) & Mean IoU \\
\midrule
 & 1--3    & 86 & 53.1 & 49.0 & \textbf{62.9} & 59.4 \\
 Harborview          & 4--10   & 32 & 48.3 & 46.3 & \textbf{52.2} & 55.5 \\
           & $>$10   & 12 & 42.6 & 46.4 & \textbf{51.9} & 51.9 \\
\midrule
      & 1--3    & 341 & 42.1 & 45.7 & \textbf{53.2} & 52.1 \\
SNUBH            & 4--10   & 124 & 30.1 & 37.7 & \textbf{46.0} & 45.0 \\
           & $>$10   & 97  & 11.0 & 15.2 & \textbf{17.7} & 24.8 \\
\bottomrule
\end{tabular}
}
\end{table}

Table~\ref{tab:gap_recovery} isolates the regime LPT targets: frames where the shared YOLO detector fails. We define a \emph{gap} as a maximal run of consecutive labelled frames in which YOLO scores IoU$<$0.5, and stratify gaps by length. For each method we report \emph{recovery}, the fraction of gap frames it re-localizes to IoU$\ge$0.5. Because gaps are defined by the detector's own failures, the detector recovers none of them by construction; any recovery comes from temporal propagation. Three observations support the role of local point tracking in temporal recovery. First, LPT recovers more detector-missed frames than CL-MC in every length bucket on both datasets (e.g.\ $62.9$ vs.\ $53.1$ on short Harborview gaps, $46.0$ vs.\ $30.1$ on medium SNUBH gaps). Second, recovery remains competitive as gaps lengthen: on Harborview LPT stays near $52\%$ recovery from medium to long gaps, while CL-MC falls to $42.6\%$ on long gaps; on SNUBH, both methods degrade, but LPT remains ahead. Third, the stateless box snap not only tightens boxes but also rescues borderline recoveries: comparing LPT~(w/o box snap) to LPT~(default), snapping a loose point-derived box to the SAM2 mask pushes a number of just-below-threshold frames over IoU$\,0.5$ (e.g.\ $37.7\rightarrow46.0$ on medium SNUBH gaps), consistent with its role in Sec.~\ref{subsec:selection_refine}.

\subsection{Ablation study}
\label{sec:ablation}

\begin{table*}[!tb]
\centering
\caption{Module ablation and sanity controls. Rows toggle temporal point tracking (\emph{CoTr.}), detector re-anchoring, deterministic candidate scoring, and final SAM2 box snap. \textbf{Bold} = best, \underline{underline} = second.}
\label{tab:abl_modules}
\scriptsize
\setlength{\tabcolsep}{4pt}
\renewcommand{\arraystretch}{0.98}
\resizebox{\linewidth}{!}{%
\begin{tabular}{cccc cccc cccc}
\toprule
\multicolumn{4}{c}{Components} & \multicolumn{4}{c}{\textsc{Harborview}} & \multicolumn{4}{c}{\textsc{SNUBH}} \\
\cmidrule(lr){1-4}\cmidrule(lr){5-8}\cmidrule(lr){9-12}
CoTr. & Re-anc. & Score & Snap & mAP$_{50}\uparrow$ & mAP$_{50:95}\uparrow$ & AUC$\uparrow$ & MR$\downarrow$ & mAP$_{50}\uparrow$ & mAP$_{50:95}\uparrow$ & AUC$\uparrow$ & MR$\downarrow$ \\
\midrule
        &           &        &           & 81.65 & 54.68 & 74.84 & 10.58 & 54.20 & 29.27 & 47.27 & 41.88 \\
        &           &        & \checkmark & 83.64 & 65.07 & 81.72 & 8.13  & 56.91 & 41.82 & 55.71 & 38.68 \\
\checkmark & \checkmark & \checkmark &        & \underline{88.32} & 57.04 & 78.48 & 6.36  & 59.96 & 30.31 & 54.91 & 34.13 \\
\checkmark & \checkmark &           & \checkmark & 85.26 & 65.87 & \underline{84.00} & \underline{5.72} & \textbf{61.04} & \textbf{43.97} & \underline{61.42} & \underline{33.42} \\
\checkmark &           & \checkmark & \checkmark & 88.10 & \underline{67.58} & 82.38 & 7.39 & 60.46 & \underline{42.54} & 60.40 & 34.89 \\
\midrule
\checkmark & \checkmark & \checkmark & \checkmark & \textbf{89.06} & \textbf{67.87} & \textbf{84.06} & \textbf{5.57} & \underline{61.02} & 42.52 & \textbf{61.61} & \textbf{33.12} \\
\bottomrule
\end{tabular}%
}
\end{table*}

Table~\ref{tab:abl_modules} separates the two main effects. Adding SAM2 refinement to YOLO improves strict localization but barely fixes temporal misses (MR $10.58\!\to\!8.13$ on Harborview, $41.88\!\to\!38.68$ on SNUBH), whereas local point recovery alone lowers MR to $6.36$ and $34.13$. Combining both yields the best Harborview mAP$_{50}$/mAP$_{50:95}$ and the best SNUBH AUC/MR, supporting the ``localize first, snap second'' decomposition: CoTracker owns temporal recovery, stateless SAM2 owns boundary tightness.

The component-removal rows show which choices are conservative rather than universally dominant. Removing the candidate-scoring rule costs $3.80$ Harborview mAP$_{50}$, so we treat it as a target-discovery safeguard rather than a dominant module. Removing detector re-anchoring is harmful on both datasets, raising MR by $1.82$ (Harborview) and $1.77$ (SNUBH).

Since the box snap re-derives the final box and can hide upstream differences, we also probe the point-to-box geometry with the snap \emph{disabled} (Table~\ref{tab:component_verify}). Across the full $2\times3$ grid of point source (SAM mask vs.\ grid) and recovery (quantile, affine, blend), mAP$_{50}$ varies within about $0.6$ points on either dataset. This parity indicates the main effect comes from using local point trajectories as the temporal state, not from finely tuning the point-to-box estimator; we default to SAM-mask initialization with the visibility-weighted blend.

\begin{table}[!htb]
\centering
\caption{\textbf{Point-sampling and box-recovery choices.} All rows use the full pre-snap pipeline, isolating query-point source and point-to-box recovery. The main experiments use SAM-mask-supported query initialization with the visibility-weighted blend; regular-grid sampling is included as an ablation. mAP$_{50}$ stays within about $0.6$ points across variants.}
\label{tab:component_verify}
\scriptsize
\setlength{\tabcolsep}{4pt}
\renewcommand{\arraystretch}{0.9}
\begin{tabular}{l l cc cc}
\toprule
Point source & Recovery & \multicolumn{2}{c}{Harborview} & \multicolumn{2}{c}{SNUBH} \\
\cmidrule(lr){3-4}\cmidrule(lr){5-6}
 & & mAP$_{50}$ & mAP$_{50:95}$ & mAP$_{50}$ & mAP$_{50:95}$ \\
\midrule
SAM mask & blend & 88.32 & 57.04 & 59.96 & 30.31 \\
SAM mask & quantile & 88.08 & 57.05 & 60.15 & 30.54 \\
SAM mask & affine & 87.93 & 56.86 & 59.68 & 30.23 \\
grid & blend & 87.71 & 56.76 & 59.72 & 30.17 \\
grid & quantile & 87.97 & 56.87 & 59.91 & 30.39 \\
grid & affine & 87.74 & 56.78 & 59.55 & 30.12 \\
\bottomrule
\end{tabular}
\end{table}

\subsection{Runtime / Efficiency}
\label{sec:exp_runtime}

\begin{table}[!htb]
\centering
\caption{Runtime/efficiency at 1440$\times$1080 on a single NVIDIA H100 (80\,GB) GPU, tested end-to-end. Accuracy columns reuse Harborview values from Table~\ref{tab:main}.}
\label{tab:runtime}
\scriptsize
\setlength{\tabcolsep}{4pt}
\renewcommand{\arraystretch}{0.9}
\resizebox{\linewidth}{!}{%
\begin{tabular}{l cc cc}
\toprule
Method & ms/frame $\downarrow$ & FPS $\uparrow$ & HV mAP$_{50}$ $\uparrow$ & HV mAP$_{50:95}$ $\uparrow$ \\
\midrule
YOLO only           & 12.9 & 77.5 & 81.65 & 54.68 \\
CL-MC               & 51.5 & 19.4 & 84.32 & 50.95 \\
\textbf{LPT} (w/o box snap)  & 19.4 & 51.4 & \underline{88.32} & 57.04 \\
\textbf{LPT} (default)  & 70.0 & 14.3 & \textbf{89.06} & \textbf{67.87} \\
\bottomrule
\end{tabular}
}
\end{table}

The runtime comparison in Table~\ref{tab:runtime} reports end-to-end runtime and FPS. LPT (w/o box snap) keeps the detector, local point propagation, and deterministic selector, but omits the final SAM2 box snap; it runs at 51.4 FPS while retaining strong mAP$_{50}$ and a low MR in Table~\ref{tab:main}. LPT (default) adds the stateless box snap and runs at 14.3 FPS, trading speed for much higher mAP$_{50:95}$, the expected cost of snapping a tight boundary on every emitted frame.

\subsection{Discussion}

Across the main results, detector-gap analysis, and ablations (Tables~\ref{tab:main}, \ref{tab:gap_recovery}, and~\ref{tab:abl_modules}), LPT's main advantage comes from changing the temporal representation rather than from adding another stateful tracker. Local point trajectories recover target location under motion blur, occlusion, and deformation, while the stateless mask prompt is most useful after the target has already been localized. The two modules therefore address different failure modes of emergency video laryngoscopy.

\paragraph{Limitations.} This work focuses on localizing a single clinically dominant target, the vocal-cord/glottic region. Extending the same local-point formulation to multi-object surgical tracking\cite{nwoye2019weakly}, where multiple instruments or anatomical structures may enter, leave, overlap, and require persistent identities, remains an important direction for future work\cite{zia2021surgical}. Recovery during long detector gaps also remains limited (Table~\ref{tab:gap_recovery}). In addition, LPT (default) performs a SAM2 box snap on each emitted frame and is therefore slower than LPT (w/o box snap). For many clinical-assistance scenarios, however, a coarse but reliable target location is sufficient for view guidance or downstream state estimation; in such settings the variant without box snap provides a more practical real-time operating point, while LPT (default) is better suited when strict boundary accuracy is required. Finally, we evaluate localization accuracy only; the effect of LPT on clinical outcomes such as intubation success, and its agreement with clinicians~\cite{jian2026equal}, remain to be studied.

\section{Conclusion}

We presented LPT, a local point-tracking framework for vocal cord localization in emergency video laryngoscopy. Instead of propagating a whole-region memory, LPT tracks SAM-initialized local points between detector anchors and uses SAM2 for stateless point initialization and final box snapping. This separation improves temporal recovery while preserving tight spatial localization, providing a simple alternative to stateful region-memory tracking in visually ambiguous clinical video.

{\small\noindent\textbf{Acknowledgements.} This work was funded by the Kren Engineering-based Medicine initiative (0925-0001 and 0925-0002), University of Washington School of Medicine, Seattle, WA. This work was also partially supported by the National Science and Technology Council (NSTC), Taiwan, under Grant NSTC 114-2634-F-006-002. We thank the National Center for High-Performance Computing (NCHC), National Institutes of Applied Research (NIAR), Taiwan, for providing computational and storage resources.\par}

{\small
\bibliographystyle{unsrtnat}
\bibliography{ref}
}

\end{document}